\pdfoutput=1
\documentclass[11pt]{article}

\usepackage{ACL2023}

\usepackage{times}
\usepackage{latexsym}
\usepackage{booktabs}
\usepackage{longtable}
\usepackage{multirow}
\usepackage{graphicx}
\usepackage{graphics}
\usepackage{caption}
\usepackage{subcaption}
\usepackage{xcolor}
\usepackage{multicol}

\usepackage[T1]{fontenc}
\usepackage[utf8]{inputenc}

\usepackage{microtype}

\usepackage{inconsolata}

\title{Automatic Readability Assessment for Closely Related Languages}

\author{Joseph Marvin Imperial$^{\Omega,\Lambda}$~\;~Ekaterina Kochmar$^{\Upsilon}$ \\
 $^{\Omega}$National University, Philippines~\;~ $^{\Lambda}$University of Bath, UK \\
 $^{\Upsilon}$MBZUAI, UAE\\
\texttt{\href{mailto:jmri20@bath.ac.uk}{jmri20@bath.ac.uk}}~\;~\texttt{\href{mailto:ekaterina.kochmar@mbzuai.ac.ae}{ekaterina.kochmar@mbzuai.ac.ae}}
 }

\begin{document}
\maketitle
\begin{abstract}
In recent years, the main focus of research on automatic readability assessment (ARA) has shifted towards using expensive deep learning-based methods with the primary goal of increasing models' accuracy. This, however, is rarely applicable for low-resource languages where traditional handcrafted features are still widely used due to the lack of existing NLP tools to extract deeper linguistic representations. In this work, we take a step back from the technical component and focus on how linguistic aspects such as {\em mutual intelligibility} or {\em degree of language relatedness} can improve ARA in a low-resource setting. We collect short stories written in three languages in the Philippines -- Tagalog, Bikol, and Cebuano -- to train readability assessment models and explore the interaction of data and features in various cross-lingual setups. Our results show that the inclusion of {\textsc{CrossNGO}}, a novel specialized feature exploiting n-gram overlap applied to languages with high mutual intelligibility, significantly improves the performance of ARA models compared to the use of off-the-shelf large multilingual language models alone. Consequently, when both linguistic representations are combined, we achieve state-of-the-art results for Tagalog and Cebuano, and baseline scores for ARA in Bikol.

We release our data and code at \texttt{\href{https://github.com/imperialite/ara-close-lang}{github.com/imperialite/ara-close-lang}}
\end{abstract}

\section{Introduction}
Automatic readability assessment (ARA) is the task that aims to approximate the difficulty level of a piece of literary material using computer-aided tools. The need for such application arises from challenges related to the misalignment of difficulty labels when humans with various domain expertise provide annotations, as well as to the difficulty of manual extraction of complex text-based features \cite{deutsch-etal-2020-linguistic}. At the same time, readability assessment tools often use different definitions of complexity levels based on (a) age level \cite{vajjala2012improving,xia-etal-2016-text}, (b) grade level \cite{imperial2020exploring,imperial2021diverse}, or on established  frameworks such as (c) the Common European Framework of Reference for Languages (CEFR)\footnote{\url{https://www.coe.int/en/web/common-european-framework-reference-languages/level-descriptions}}~\cite{franccois2012ai,pilan2016readable,xia-etal-2016-text,reynolds-2016-insights,vajjala-rama-2018-experiments}. 

In recent years, deep learning methods and large language models (LLMs) have gained popularity in the research community. Often studies using these methodologies focus primarily on improving the performance across various metrics. This is particularly manifest in ARA research in languages with a high number of accessible and publicly-available readability corpora such as English~\cite{heilman-etal-2008-analysis,flor-etal-2013-lexical,vajjala-lucic-2018-onestopenglish} and German \cite{hancke-etal-2012-readability,weiss-etal-2021-using,weiss-meurers-2022-assessing} to name a few. At the same time, existing studies focusing on low-resource languages such as Cebuano \cite{imperial-etal-2022-baseline} and Bengala \cite{islam2012text,islam2014readability} are still at the stage of primarily using traditional features such as word and sentence lengths to train predictive models. 

We identify two problems that are related to the use of complex neural-based approaches: the success of such models depends on (a) whether there is enough available data to train a model using a customized deep neural network, and (b) in the case of LLMs, whether there exists an available off-the-shelf pre-trained model for a low-resource language of interest. \citet{imperial-etal-2022-baseline} have recently shown that merely integrating extracted embeddings from a multilingual BERT model as features for Cebuano, a low-resource Philippine language, \textit{does not outperform} models trained with orthographic features such as syllable patterns customized for the language. These challenges provide motivation for researchers to further explore methods that do not rely on the availability of large amounts of data or complex pre-trained models and investigate simpler, more interpretable models instead of black box architectures. 

In this paper, we take a step back and focus on the data available for low-resource Philippine languages and the features extracted from them rather than on the algorithmic aspects. Specifically, we explore a scenario where small readability corpora are available for languages that are {\em closely related} or belong to one major language family tree. To the best of our knowledge, incorporating the degree of language closeness or relatedness has not been explored before in any cross-lingual ARA setup. In this study, we make the following contributions:

\begin{enumerate}
    \item We conduct an extensive pioneer study on readability assessment in a cross-lingual setting using three closely related Philippine languages: Tagalog, Bikolano, and Cebuano.
    \item We extract various feature sets ranging from linguistically motivated  to neural embeddings, and empirically evaluate how they affect the performance of readability models in a singular, pairwise, and full cross-lingual setup.
    \item We introduce cross-lingual  Character N-gram Overlap ({\sc CrossNGO}), a novel feature applicable to readability assessment in closely related languages.
    \item We also introduce and release a new readability corpus for Bikolano, one of the major languages in the Philippines.
    \item Finally, we set a baseline for ARA in Bikol and report state-of-the-art results for Tagalog and Cebuano. 
\end{enumerate}

\section{Background}
\label{sec:prelim}

\subsection{The Philippine Linguistic Profile}
\begin{figure}[!htbp]
    \centering
    \includegraphics[width=.52\textwidth,trim={5.5cm 1.5cm 2cm 1cm}, clip]{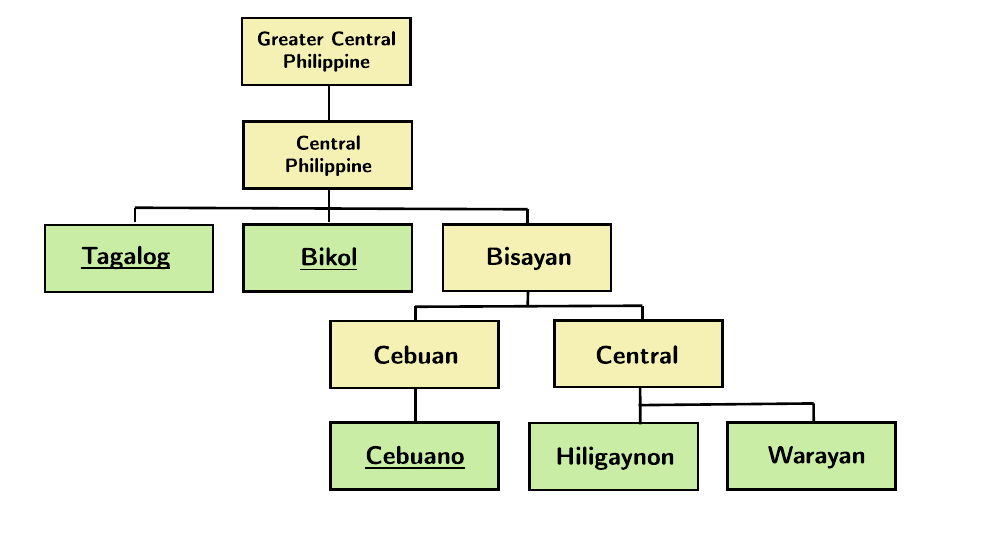}
    \caption{The central subgroup of the Philippine language family tree highlighting the origins of Tagalog, Bikol, and Cebuano.}
    \label{fig:language_tree}
\end{figure}

The Philippines is a linguistically diverse country in Southeast Asia (SEA) with over $180$ languages spoken by over $100$ million people. Languages in the Philippines can be best described as morphologically rich due to their free-word order structures and high number of possible inflections, full and partial duplications, and compound words \cite{go-nocon-2017-using}. In addition, following  lexicostatistical studies, languages are divided into two subgroups, {\em northern} and {\em central}, wherein the major languages Ilokano, Pangasinan, and Kapampangan belong to the northern subgroup, and Tagalog, Bikol, Hiligaynon, and Cebuano are allocated to the central subgroup \cite{walton1979philippine,constantino1998current}. Figure~\ref{fig:language_tree} illustrates the central subgroup of the Philippine language family tree.

%%
%% - LANGUAGE FAMILY TREE
%%

In this study, our readability experiments focus on three major Philippine languages, {\em Tagalog}, {\em Cebuano}, and {\em Bikol}, which we refer to further in the paper with their corresponding ISO-639-2 language codes as \textsc{tgl}, \textsc{ceb}, and \textsc{bcl}, respectively.
%% linguistic history and stats 
%% linguistic structure, definition of closely related, morphologically rich, %% bicol, tagalog, cebuano, family tree structure

\subsection{Mutual Intelligibility}
Preliminary linguistic profiling studies of the main Philippine languages such as by \citet{mcfarland2004philippine} show that Tagalog, Bikol, and Cebuano are more closely related to one another than any languages in the northern family tree. A language's closeness or its \textit{degree of relatedness} to another language from the same family (sub)tree is commonly referred to as {\em mutual intelligibility} \cite{bloomfield1926set}. Such similarities can be seen across multiple aspects, including, for example (a) syllable patterns where all three languages have similar three case-marking particles – \textit{ang} (En: \textit{the}), \textit{ng} (En: \textit{of}), and \textit{sa} (En:  \textit{at}) for Bikol and Tagalog, and \textit{ug} instead of \textit{sa} for Cebuano; and (b) shared words, e.g. \textit{mata} (En: \textit{eye}) and \textit{tubig} (En: \textit{water}). 

For languages belonging to one greater subgroup in the case of Central Philippine for Tagalog, Bikol, and Cebuano, showing stronger quantitative evidence of mutual intelligibility may provide additional proof that these languages are indeed, at some level, closely related to each other. Thus, to contribute towards further understanding of mutual intelligibility in the Philippines language space, we apply two linguistic similarity-based measures using character n-gram overlap and genetic distance which we discuss in the sections below.\newline

%This finding encourages research efforts on measuring the \textbf{degree of relatedness} between any two languages within this group via manual or computational analysis in text and speech.

%In this work, we use short stories as the data for readability analysis. 

%%
%% NGRAM OVERLAP TABLE
%%
\begin{table}[!htbp]

\begin{subtable}[h]{\linewidth}
    \small
    \centering
    \begin{tabular}{@{}lllll@{}} \toprule
                 & \textbf{\textsc{TGL}} & \textbf{\textsc{BCL}} & \textbf{\textsc{CEB}} & \textbf{\textsc{ENG}} \\\midrule
                \textbf{\textsc{TGL}} & 1.000            &    0.810            &     0.812             &  0.270                \\
                \textbf{\textsc{BCL}}   & 0.810            & 1.000          &     0.789             &   0.263               \\
                \textbf{\textsc{CEB}} & 0.812            & 0.789          & 1.000            &  0.213                \\
                \textbf{\textsc{ENG}} & 0.270            & 0.263          & 0.213            & 1.000   \\
                \bottomrule
                \end{tabular}
                \caption{Bigram Character Overlap}
                \label{tab:mutual_bigram}
\end{subtable}
\newline
\vspace*{0.4cm}
\newline
\begin{subtable}[h]{\linewidth}
    \small
    \centering
    \begin{tabular}{@{}lllll@{}} \toprule
                 & \textbf{\textsc{TGL}} & \textbf{\textsc{BCL}} & \textbf{\textsc{CEB}} & \textbf{\textsc{ENG}} \\\midrule
                \textbf{\textsc{TGL}} & 1.000            &    0.588            &  0.628                &   0.121               \\
                \textbf{\textsc{BCL}}   & 0.588            & 1.000          &  0.533                &   0.144               \\
                \textbf{\textsc{CEB}} & 0.628            & 0.533          & 1.000            &   0.090               \\
                \textbf{\textsc{ENG}} & 0.121            & 0.144          & 0.090           & 1.000   \\
                \bottomrule
                \end{tabular}
                \caption{Trigram Character Overlap}
                \label{tab:mutual_trigram}
\end{subtable}

\caption{Mutual Intelligibility using Bigram and Trigram Character N-Gram Overlap}
\label{tab:mutual_charngram}
\end{table}

\noindent \textbf{Character N-Gram Overlap.} For our first measure, we use the overlap in character bigrams and trigrams for every pair from the selected set of languages. To do this, we simply extract and rank the top occurring character bigrams and trigrams for a given language and calculate the Rank-Biased Overlap (RBO)\footnote{\url{https://github.com/changyaochen/rbo}} 
\cite{webber2010similarity}. RBO provides a measure of similarity between two lists while preserving the ranking. We also add English ({\sc eng}) as an unrelated control language not belonging to the Philippine family tree for comparison. We use the CommonCore readability dataset \cite{flor-etal-2013-lexical} for English as it also has three readability levels, and the level distribution is the most similar to the dataset of the three Philippine languages. Further information on the datasets in Tagalog, Bikol, and Cebuano can be found in Section \ref{sec:readability}. For all languages, we extract the top $25\%$ of the most frequently occurring bigrams and trigrams for analysis. The top $40$ most frequent bigrams and trigrams can be found in the Appendix.

Table~\ref{tab:mutual_charngram} presents character overlap for bigrams and trigrams in a pairwise manner. These results show that all three Philippine languages have character overlap greater than $75\%$ for bigrams among themselves while overlap with English is below $27\%$. This pattern is observed again in trigrams with the overlap levels of $53.3\%$ to $62.8\%$ between Tagalog, Bikol, and Cebuano and those below $15\%$ for English. These ranges of mutual intelligibility values for bigram and trigram overlap serve as an estimate of the degree of relatedness between the three Philippine languages, with the values for English serving as a baseline for an unrelated language.\newline

\noindent \textbf{Genetic Distance.} As a secondary measure of mutual intelligibility, we calculate the genetic distance score \cite{beau-crabbe-2022-impact} for each pair of languages studied in this work. Similar to the character n-gram overlap analysis, we add English for comparison purposes. Genetic distance \cite{beaufils2020stochastic} is an automatic measure for quantifying the distance between two languages without the need for human judgments. This metric requires a list of words and their equivalent translations for any two languages of interest and calculates the number of exact consonant matches using the following formula:

\begin{equation}
\mathrm{Genetic Distance} = 100-(\frac{match({l}_{1},{l}_{2})}{n})
\end{equation}

\noindent where ${l}_{1}$ and ${l}_{2}$ are a pair of languages, $n$ is the total number of words for analysis (usually $100$), and $match(\cdot)$ is a function for extracting the consonant patterns for each word from the list as described in \citet{beaufils2020stochastic}. The metric is measured as a distance; thus, the values closer to $100$ denote higher dissimilarity or non-relatedness.

%%
%% GENETIC DISTANCE TABLE
%%
\begin{table}[!htbp]
\small
\centering
\begin{tabular}{@{}lrrrr@{}} \toprule
 & \textbf{\textsc{TGL}} & \textbf{\textsc{BCL}} & \textbf{\textsc{CEB}} & \textbf{\textsc{ENG}} \\\midrule
\textbf{\textsc{TGL}} & 0.000            &  37.083              &  24.846                &  95.690                \\
\textbf{\textsc{BCL}} & 37.083            & 0.000          & 31.933                 & 70.735                 \\
\textbf{\textsc{CEB}} & 24.846            & 31.933          & 0.000            &  90.970                 \\
\textbf{\textsc{ENG}} & 95.690            & 70.735          & 90.970            & 0.000   \\
\bottomrule
\end{tabular}

\vspace*{0.4cm}

\begin{tabular}{@{}ll@{}} \toprule
\textbf{Range} & \textbf{Meaning} \\ \midrule
{\color[HTML]{006600} \textbf{Between 1 and 30}}   & Highly related languages        \\
{\color[HTML]{009901} \textbf{Between 30 and 50}}  & Related languages               \\
{\color[HTML]{32CB00} \textbf{Between 50 and 70}}  & Remotely related languages      \\
{\color[HTML]{FFC702} \textbf{Between 70 and 78}}  & Very remotely related languages \\
{\color[HTML]{FE0000} \textbf{Between 78 and 100}} & No recognizable relationship   \\\bottomrule
\end{tabular}

\caption{Mutual Intelligibility using Genetic Distance with Mapping from \citet{beaufils2020stochastic}.}
\label{tab:genetic_distance}
\end{table}

Table~\ref{tab:genetic_distance} shows the calculated genetic distance scores for each pair of languages including English. The mapping provided in the table is the prescribed guide from \citet{beaufils2020stochastic}. Judging by these results, the Philippine languages have genetic distance scores within the \textcolor[HTML]{009901}{\textbf{related}} and \textcolor[HTML]{006600}{\textbf{highly related languages}} range with the Tagalog–Cebuano pair showing the closest language distance of $24.846$. Meanwhile, genetic distance scores between all considered Philippine languages and English fall within the \textcolor[HTML]{FFC702}{\textbf{very remotely related}} to \textcolor[HTML]{FE0000}{\textbf{no recognizable relationship}} categories, with the Tagalog–English pair showing the highest distance from each other. Similar to the character n-gram overlap, these results strengthen our initial observation and provide empirical evidence for mutual intelligibility between Tagalog, Bikol, and Cebuano languages which, beyond this study, may also be used in future linguistic research.

\section{Readability Corpora in Philippine Languages}
\label{sec:readability}
We have compiled open source readability datasets for Tagalog, Cebuano, and Bikol from online library websites and repositories. Each data instance in this study is a fictional short story. Table~\ref{data} shows the statistical breakdown and additional information on the levels in each readability dataset across different languages. \newline

\noindent \textbf{Tagalog and Cebuano.} Datasets in these languages have already been used in previous research, including \citet{imperial2019developing,imperial2020exploring,imperial-2021-bert,imperial2021diverse,imperial-etal-2022-baseline}. We use the same datasets as in previous research and incorporate them into this study for comparison. For Tagalog, we have assembled $265$ instances of children's fictional stories from Adarna House\footnote{\url{https://adarna.com.ph/}} and the Department of Education (DepED)\footnote{\url{https://lrmds.deped.gov.ph/}}. For Cebuano, we use the dataset collected  by \citet{imperial-etal-2022-baseline} from Let's Read Asia\footnote{\url{https://www.letsreadasia.org/}} and Bloom Library\footnote{\url{https://bloomlibrary.org/}}, which were funded by the Summer Institute of Linguistics (SIL International) and BookLabs to make literary materials in multiple languages available to the public. \newline

\noindent \textbf{Bikol.} There are no pre-compiled datasets available for readability assessment in Bikol yet. For this, we collected all available Bikol short stories from Let's Read Asia and Bloom Library totaling $150$ instances split into $68$, $27$, and $55$ for levels $1$ to $3$ respectively. \newline

\noindent All collected data for this study follows the standard leveling scheme for early-grade learners or the first three grades from the K-12 Basic Curriculum in the Philippines.\footnote{\url{https://www.deped.gov.ph/k-to-12/about/k-to-12-basic-education-curriculum/}} Each instance has been annotated by experts with a level from $1$ to $3$ as seen in Table~\ref{data}. We use these annotations as target labels in our experiments. Finally, all datasets used in this study can be manually downloaded from their respective websites (see footnotes for links) under the Creative Commons BY 4.0 license.

%which allows redistribution and transformation in any format for both commercial and research purposes.

%%
%% DATA TABLE
%%
\begin{table*}[!htbp]
\small
\centering
\begin{tabular}{@{}lccccc@{}}
\toprule
\multicolumn{1}{c}{\textbf{Source}} &
  \textbf{Language} &
  \textbf{Level} &
  \textbf{Doc Count} &
  \textbf{Sent Count} &
  \textbf{Vocab} \\ \midrule
\multirow{3}{*}{\begin{tabular}[c]{@{}l@{}} Adarna and DepED\end{tabular}} &
  \multirow{3}{*}{\textsc{TGL} (265)} &
  L1 &
  72 &
  2774 &
  4027 \\
 &  & L2 & 96  & 4520  & 7285  \\
 &  & L3 & 97  & 10957 & 12130 \\ \midrule
\multirow{3}{*}{\begin{tabular}[c]{@{}l@{}}Let's Read Asia and \\ Bloom Library\end{tabular}} &
  \multirow{3}{*}{\textsc{BCL} (150)} &
  L1 &
  68 &
  1578 &
  2674 \\
 &  & L2 & 27  & 1144  & 2009  \\
 &  & L3 & 55  & 3347  & 5509  \\ \midrule
\multirow{3}{*}{\begin{tabular}[c]{@{}l@{}}Let's Read Asia and \\ Bloom Library\end{tabular}} &
  \multirow{3}{*}{\textsc{CBL} (349)} &
  L1 &
  167 &
  1173 &
  2184 \\
 &  & L2 & 100 & 2803  & 4003  \\
 &  & L3 & 82  & 3794  & 6115  \\ \bottomrule
\end{tabular}
\caption{Statistics on the readability corpora in Tagalog, Cebuano, and Bikol used in this study. The numbers in the brackets provided in the second column are the total number of documents per language broken down in the third and fourth columns per grade level. \textbf{Doc Count} and \textbf{Sent Count} denote the number of short story instances and the number of sentences per story. \textbf{Vocab} is the size of the vocabulary  or of the accumulated unique word lists per level.}
\label{data}
\end{table*}

\section{Experimental Setup}
\label{sec:setup}

\subsection{ML Setup}
In this study, our primary focus is on the depth of analysis of the traditional and neural features used in a cross-lingual setting applied to closely related languages. Thus, we use a vanilla Random Forest model which has been previously shown to be the best-performing monolingual-trained model for ARA in Tagalog and Cebuano~\cite{imperial2021diverse, imperial-etal-2022-baseline}. We leave the technical breadth of exploring other supervised algorithms to future work. 

We use a stratified $k$-fold approach with $k$$=$$5$ to have well-represented samples per class for a small-dataset scenario used in this study. We report accuracy as the main evaluation metric across all experiments for the ease of performance comparison with previous work (see Section \ref{sec:results}). We use WEKA $3.8$ \cite{witten1999weka}\footnote{\url{https://www.cs.waikato.ac.nz/ml/weka/}} for all our modeling and evaluation and set hyperparameters of the Random Forest algorithm to their default values as listed in the Appendix.

\subsection{Linguistic Features}
We extract and consider a wide variety of features inspired by: (a) handcrafted predictors from previous work, (b) representations from a multilingual Transformer-based model (mBERT), and (c) \textsc{CrossNGO}, a novel feature applicable to readability assessment in closely related languages. We discuss each feature group below. \newline

%% cite sources that have used these features

\noindent \textbf{Traditional Handcrafted Features (\textsc{TRAD}).} We integrate available traditional surface-based and syllable pattern-based features in this study as predictors of text complexity. These features have been widely used in previous research on ARA in Tagalog and Cebuano \cite{imperial2020exploring,imperial-etal-2022-baseline}. For Bikol, this is the first-ever study to develop a readability assessment model. In the case of low resource languages similar to those used in this study, these predictors are still the go-to features in ARA and have been empirically proven effective for Tagalog and Cebuano \cite{imperial-ong-2021-microscope}. We have extracted a total of $18$ traditional features for each language, including:

\begin{enumerate}
    \item The total number of words, phrases, and sentences (3).
    \item Average word length, sentence length, and the number of syllables per word (3).
    \item The total number of polysyllable words of more than $5$ syllables (1).
    \item Density of consonant clusters or frequency of consonants without intervening vowels in a word (e.g. Tagalog: \textit{sa\underline{str}e}, En: {\em dressmaker}) (1).
    \item Densities of syllable patterns using the following templates \{v, cv, vc, cvc, vcc, ccv, cvcc, ccvc, ccvcc, ccvccc\}, where v and c are vowels and consonants respectively (10).
\end{enumerate}

\noindent \textbf{Multilingual Neural Embeddings  (mBERT).} In addition to the surface-based features, we explore contextual representations from a multilingual Transformer-based large language model via mBERT \cite{devlin-etal-2019-bert}. Previous research on probing BERT has shown convincing evidence that various types of linguistic information (e.g. semantic and syntactic knowledge) are distributed within its twelve layers
\cite{tenney-etal-2019-bert,rogers-etal-2020-primer}. Applying this to ARA, \citet{imperial-2021-bert} showed that BERT embeddings could act as a \textit{substitute} feature set for lower-resource languages such as Filipino, for which NLP tools like POS taggers are lacking. 

For this study, we specifically chose mBERT as this particular model has been trained using Wikipedia data in $104$ different languages including Tagalog and Cebuano. Bikol is not included in any available off-the-shelf Transformer-based language models due to extremely limited online resources not large enough for training. Nonetheless, we still used the representations provided by mBERT noting its high intelligibility with Tagalog and Cebuano. Feature-wise, we use the mean-pooled representations of the entire twelve layers of mBERT via the \texttt{sentence-transformers} library \cite{reimers-gurevych-2019-sentence}. Each instance in our readability data has an mBERT embedding representation of $768$ dimensions. \newline

\noindent \textbf{Cross-lingual Character N-Gram Overlap (\textsc{CrossNGO}).} 
N-gram overlap has been used previously in various NLP tasks applied to Philippine language data such as language identification \cite{oco2013measuring,cruz2016phoneme}, spell checking and correction \cite{cheng2007spellchef,octaviano2017spell,go2017gramatika}, and clustering \cite{oco2013dice}. Drawing inspiration from this fact and from the quantitative evidence of mutual intelligibility between Philippine languages presented in Section~\ref{sec:prelim}, we posit that a new feature designed specifically for closely related language data might improve the performance of the readability assessment models. Thus, we introduce \textsc{CrossNGO}, which quantifies linguistic similarity using character overlap from a curated list of high-frequency n-grams within languages of high mutual intelligibility. We propose the following formula for calculating this metric:

\begin{equation}
\mathrm{CrossNGO}_{L,n} = \frac{m(L)\bigcap m(d)}{\mathrm{count}(m(d))}
\end{equation}

\noindent where $n\in \{2,3\}$ denotes bigrams and trigrams, and $m(\cdot)$ is a function that extracts unique n-grams from a document instance $d$ and compares them to a list of top n-grams from a specific language $L$. For each instance in a dataset, a vector containing three new features will be added representing the overlap between the text and the top n-grams from each of the three languages. We apply this calculation to both bigrams and trigrams using the n-gram lists for Tagalog, Bikol, and Cebuano obtained from the preliminary experiments, which results in a total of $6$ new features.

While we presented two quantitative methods of mutual intelligibility in Section ~\ref{sec:prelim}, only \textsc{CrossNGO} is applied as a metric \textit{and} a feature for this study. Staying faithful to the work of \citet{beaufils2020stochastic}, we did not use Genetic Distance to generate another set of features as it was originally developed as a language-to-language metric. Thus, we use it only as additional secondary evidence of language similarity. At the same time, we note that the proposed \textsc{CrossNGO} bears certain conceptual similarities to Genetic Distance as it measures the frequency of n-gram overlap with other languages. We perform an ablation study and demonstrate the contribution of individual feature sets in Section \ref{sec:results}.

%%
%% RESULTS
%%
\begin{table*}[!htbp]
    \scriptsize
    \centering
    \setlength\tabcolsep{5pt}
    \begin{tabular}{@{}l|cccc|cccc|cccc@{}}
    \toprule
    \multirow{3}{*}{\textbf{Model}} & \multicolumn{4}{c}{\textbf{\textsc{TGL}}} & \multicolumn{4}{|c}{\textbf{BCL}} & \multicolumn{4}{|c}{\textbf{CEB}} \\ \cmidrule{2-13}
     &
      \textbf{\begin{tabular}[c]{@{}c@{}}TRAD\end{tabular}} &
      \textbf{\begin{tabular}[c]{@{}c@{}}TRAD + \\ CrossNGO\end{tabular}} &
      \textbf{\begin{tabular}[c]{@{}c@{}}mBERT\\ Embdng\end{tabular}} &
      \textbf{ALL} &
      \textbf{\begin{tabular}[c]{@{}c@{}}TRAD\end{tabular}} &
      \textbf{\begin{tabular}[c]{@{}c@{}}TRAD + \\ CrossNGO\end{tabular}} &
      \textbf{\begin{tabular}[c]{@{}c@{}}mBERT\\ Embdng\end{tabular}} &
      \textbf{ALL} &
      \textbf{\begin{tabular}[c]{@{}c@{}}TRAD\end{tabular}} &
      \textbf{\begin{tabular}[c]{@{}c@{}}TRAD + \\ CrossNGO\end{tabular}} &
      \textbf{\begin{tabular}[c]{@{}c@{}}mBERT\\ Embdng\end{tabular}} &
      \textbf{ALL} \\ \cmidrule{1-13}
    \textbf{\textsc{TGL}}                     & 43.153 & 44.231    & 46.100   & 50.000  & 55.172 & 41.379    & 20.689    & 24.137  & 53.623 & 57.971    & 47.826    & 50.725   \\
    \textbf{\textsc{BCL}}                    & 50.000 & \underline{50.100}    & 26.921   & 23.077   & 74.620 &\underline{75.862}    & 68.965    & 69.000   & 63.768 & 62.320    & 60.869    & 66.667   \\
    \textbf{\textsc{CEB}}                   & 32.692 & 38.462    & 34.615   & 42.308  & 51.720 & 65.517    & 48.276    & 44.823   & 74.058 & \underline{78.270}    & 71.015    & 73.913  \\
    \textbf{\textsc{ENG*}}                   & 26.923 & 44.230    &  28.846  & 26.923  & 48.275 & 37.681    &  48.250   &  48.275 & 46.375 & 62.018    & 43.478    & 43.376  \\\midrule
    
    \textbf{TGL+BCL}          & 51.101 & 51.923    & 40.384    &\underline{\textbf{57.692}} & 72.441 & 69.965    & 69.000    & 68.966    & 56.521 & 60.869    & 62.318    & 69.565    \\
    \textbf{BCL+CEB}         & 48.077 & 50.000    & 42.307    & 48.076        & 68.956  & 72.414    & 75.611    & \underline{75.862}    & 74.400 & 75.362    & 75.362    & \underline{\bf{79.710}}    \\
    \textbf{CEB+TGL}        & 44.230 & 36.538    & 48.076    & 48.100         & 52.720 & 55.172    & 41.379    & 34.483   & 77.711 & 76.811    & 73.913    & 74.464  \\ \midrule
    
    \textbf{ALL}        & 50.000 & \underline{52.910} &   46.153 &   32.692 &   72.413 &79.113 &
  65.517 &   \underline{\textbf{79.328}} &   77.710 & 78.000 &   \underline{78.261} &   75.630  \\
    
    \bottomrule
    \end{tabular}
\caption{The accuracy of cross-lingual modeling per language with various iterations using different combinations of traditional and neural-based features. The \underline{underlined} values correspond to the best model for each of the three setups while the \textbf{boldfaced} values correspond to the overall highest-performing model for each language across all setups. We included English as counter-evidence only for the singular cross-lingual setup.}
\label{tab:crossmodelling_result}
\end{table*}

\section{Results and Discussion}
\label{sec:results}

Table~\ref{tab:crossmodelling_result} shows the accuracy values obtained when training Random Forest models with various combinations of feature groups for each language of interest. The experiments were divided into three setups: (a) {\em singular cross-lingual} (${l}_{1}{\rightarrow}{l}_{2}$), (b) {\em pairwise cross-lingual} ($[{l}_{1}+{l}_{2}] {\rightarrow} {l}_{3}$), and (c) {\em full cross-lingual} ($[{l}_{1}+{l}_{2}+{l}_{3}]{\rightarrow}{l}_{1}$), each corresponding to a separate  subsection of Table~\ref{tab:crossmodelling_result}. We use the term cross-lingual in this context when a model is trained with a readability corpus from a chosen language ${l}_{n}$ or a combination of languages and evaluated with a test set from another language ${l}_{m}$ as is demonstrated in Table~\ref{tab:crossmodelling_result}. Similar to our preliminary experiments (Section \ref{sec:prelim}), we include English using the CommonCore dataset as counter-evidence for comparison with closely related languages.

\subsection{Low-Resource Languages Benefit from Specialized Cross-lingual Features}
For the singular cross-lingual experiments, the effectiveness of exploiting the bigram and trigram overlap via \textsc{CrossNGO} is demonstrated by high scores for Bikol and Cebuano ($75.862$ and $78.270$) and comparable performance for Tagalog ($50.100$). Moreover, only for this setup, there is an observed trend where traditional features combined with \textsc{CrossNGO} outperform mBERT embeddings or the combination of all features for the respective language pair ${l}_{1}{\rightarrow}{l}_{2}$. For Tagalog, this results in $50.100$ vs. $26.921$ and $23.077$; for Bikol -- $75.862$ vs. $68.965$ and $69.000$; for Cebuano -- $78.270$ vs. $71.015$ and $73.913$. In terms of cross-linguality, in the case of Tagalog, using a model trained with Bikol data proves to be more effective than training with the original Tagalog data with approximately $5.8$-point difference in accuracy. However, we still recommend the Tagalog model using all features with $50.000$ accuracy since the $0.1$ difference is not a significant improvement. Consequently, this trend is not observed in the Bikol and Cebuano experiments where the best-performing models of readability assessment are trained on the data from the same language ${l}_{1}{\rightarrow}{l}_{1}$. 

To further confirm if the addition of the \textsc{CrossNGO} feature statistically improves models' performance as compared to the representations from mBERT for low-resource languages, we aggregate the scores from the TRAD+\textsc{CrossNGO} group and compare them with the scores obtained when we use mBERT embeddings only,  conducting a $t$-test. We did not include the scores using the combination of all types of features as it would confound the significance test. We achieve statistical significance at $\alpha = 0.01$ level ($p = 0.006$) which shows that using traditional handcrafted features extended with \textsc{CrossNGO} significantly improves ARA models for low-resource languages, \textit{provided} the availability of data in a closely related language in the case of non-availability of multilingual LLMs (e.g., lack of mBERT model in Bikol).

%This can be further observed via confusion matrices in Figure~\ref{fig:tag_tag_ceb}. Using a Cebuano model obtains better classification results for grades 1 and 3 but is confused between grades 2 and 3 compared to using the Tagalog model where the confusion happens in the opposite. 

%% FIG - CONFUSION MATRICES 
%\begin{figure}[!htbp]
%    \centering
    
%    \begin{subfigure}[b]{\linewidth}
%        \includegraphics[width=.45\textwidth]{VIZ/tag_tag_tradcngo.png}\hfill
%        \includegraphics[width=.45\textwidth]{VIZ/ceb_tag_tradcngo.png}\hfill
%    \end{subfigure}

%    \caption{Confusion matrices for a Tagalog (left) and Cebuano model (right) applied on a Tagalog test set.}
%    \label{fig:tag_tag_ceb}
%\end{figure}

\subsection{Inclusion of a Closely Related Language in Data Produces More Confident Predictions}
\begin{figure}[!htbp]
    \centering
    
    \begin{subfigure}[b]{\linewidth}
        \centering
        \includegraphics[width=.45\textwidth]{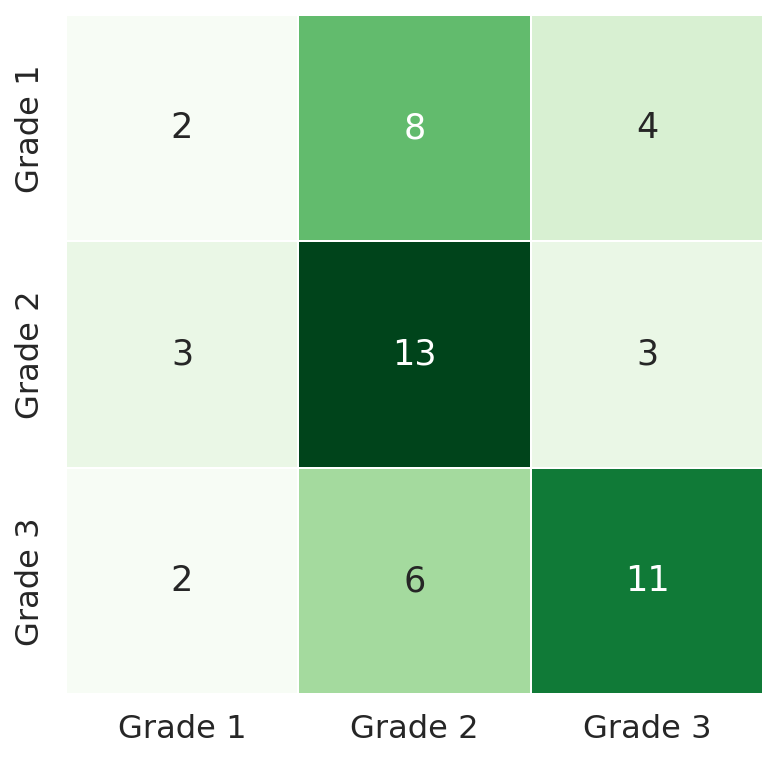}\hfill
        \includegraphics[width=.45\textwidth]{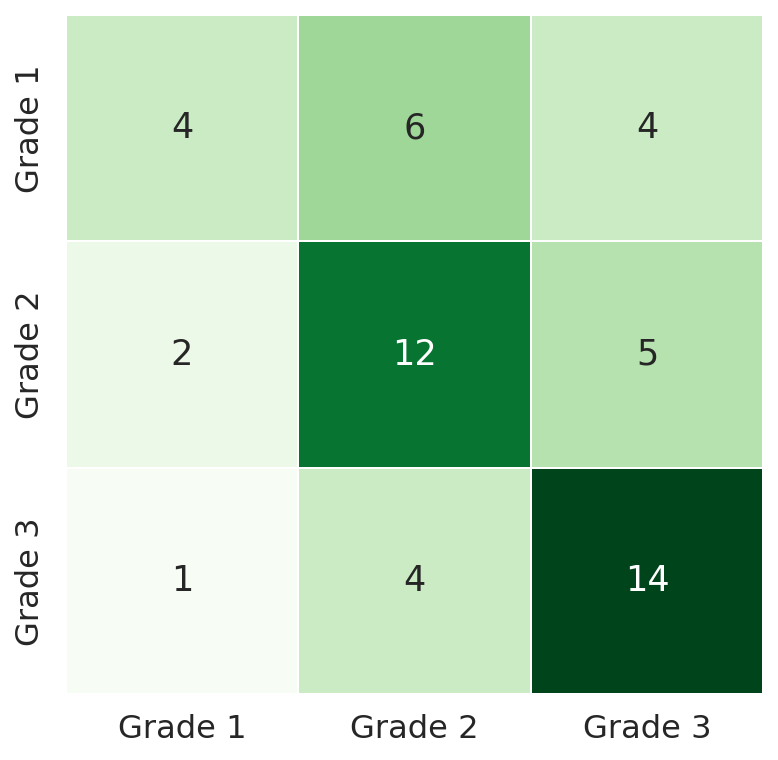}\hfill
        \caption{TGL only trained model (left) against TGL+BCL model (right) using all features for ARA in TGL.}
    \end{subfigure}
    
    \vspace*{0.4cm}
    
    \begin{subfigure}[b]{\linewidth}
        \centering
        \includegraphics[width=.45\textwidth]{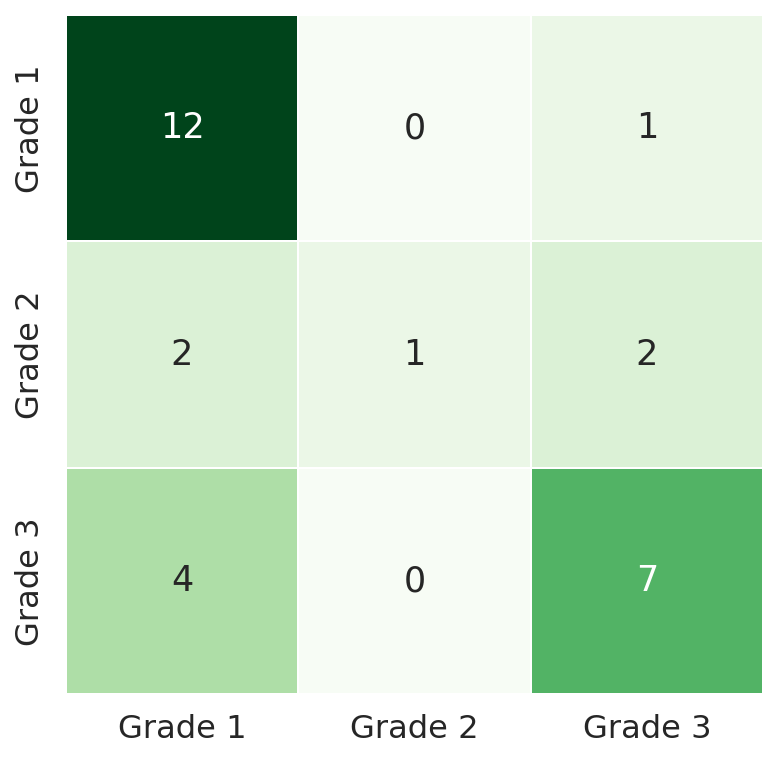}\hfill
        \includegraphics[width=.45\textwidth]{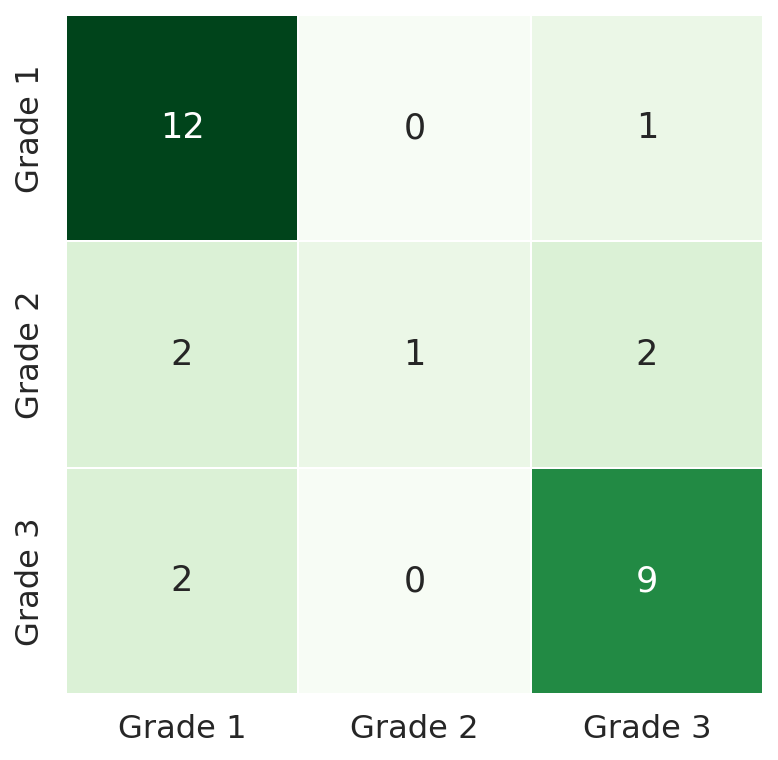}\hfill
        \caption{BCL only trained model (left) against BCL+CEB model (right) using all features for ARA in BCL.}
    \end{subfigure}

    \vspace*{0.4cm}
    
    \begin{subfigure}[b]{\linewidth}
        \centering
        \includegraphics[width=.45\textwidth]{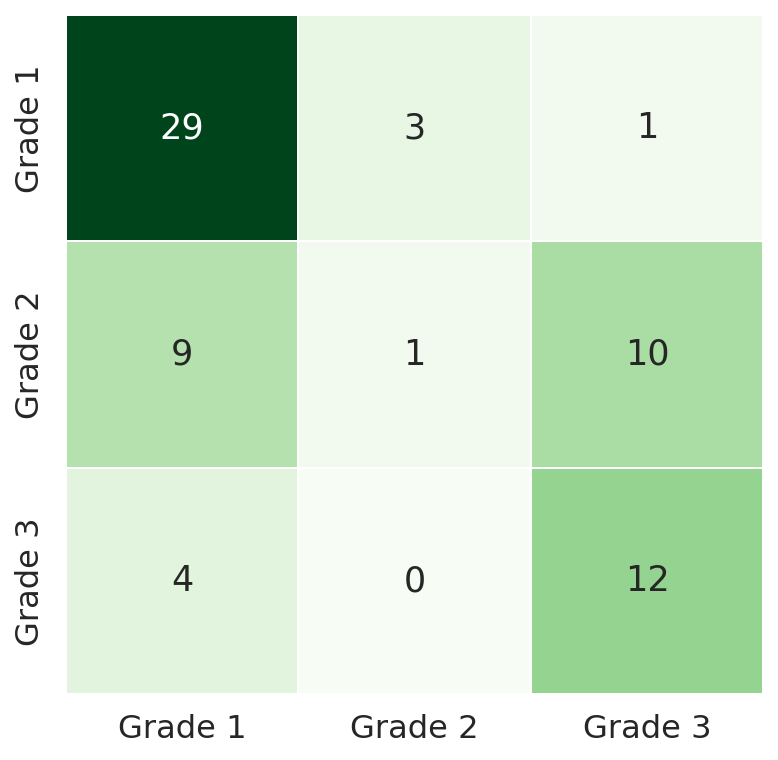}\hfill
        \includegraphics[width=.45\textwidth]{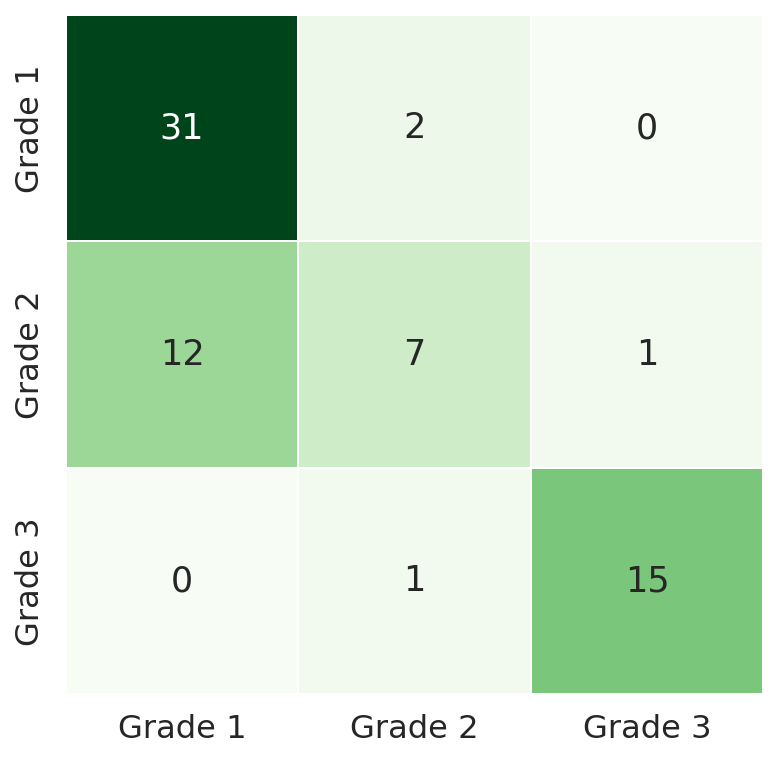}\hfill
        \caption{BCL only trained model (left) against BCL+CEB model (right) using multilingual embeddings for ARA in CEB.}
    \end{subfigure}

    \caption{Confusion matrices for pairwise cross-lingual setup for all three languages. All models using an additional language dataset for ARA achieved an improved performance with an average of $10.131$ across the board ($7.692$ for TGL, $6.862$ for BCL, and $15.841$ for CEB).}
    \label{fig:crosslingual_confusionmatrices}
\end{figure}

For pairwise cross-lingual experiments, we investigate the effect of adding a closely related language on a model's performance using confusion matrices. As the middle section of Table~\ref{tab:crossmodelling_result} demonstrates, there are three possible pairwise combinations of Tagalog, Bikol, and Cebuano tested on each individual language. As there can be numerous ways to analyze the table, we highlight the results of the cross-lingual models with the top-performing pair and their utilized feature groups and compare them to their equivalent models in the singular cross-lingual experiment. Figure~\ref{fig:crosslingual_confusionmatrices} illustrates this method of comparison for each language.

In the case of the Tagalog–Tagalog pair, most misclassifications occur between grades $1$ and $2$ in both training and test data using all features. This, in turn, is alleviated by incorporating the Bikol dataset in the training data, which reduces the level of confusion by approximately $7\%$. The inclusion of Bikol also improves classification between grades $2$ and $3$ by three instances. In the case of the Bikol test data, the same finding is observed for the combined Bikol and Cebuano model using all features, where confusion in classifying grades $1$ and $3$ is reduced by two instances. Lastly, for Cebuano, the top-performing model in the pairwise cross-lingual setup includes Bikol data and uses all features. For this model, misclassifications in predicting grade $1$ against the other two levels are reduced, and performance for predicting grade $3$ is improved. 

We further corroborate our observations that pairwise cross-lingual models outperform singular cross-lingual models by aggregating the scores from the two setups and running a $t$-test. Further to the results reported in the previous section, we observe statistically significant difference at the $\alpha = 0.01$ level $(p = 0.003)$ when pairwise cross-lingual models are compared to singular cross-lingual models. Overall, our findings provide solid empirical evidence that including a closely related language in the training data for a low-resource language significantly improves  performance.

%% FIG - CONFUSION MATRICES 

\subsection{Combining Specialized Cross-Lingual Features with Multilingual Neural Embeddings Achieves SOTA Results}
While the previous sections highlight the significant increase in performance when using traditional features with \textsc{CrossNGO} as compared to mBERT embeddings only, we now discuss results and contributions when both linguistic representations are combined. As is demonstrated in Table~\ref{tab:crossmodelling_result}, the scores obtained using the combined features applied to Tagalog and Cebuano achieve state-of-the-art results for ARA in these languages. For Tagalog, our model's accuracy of $57.692$ outperforms the SVM with $57.10$ accuracy and the Random Forest model with $46.70$ presented in \citet{imperial-2021-bert}. For Cebuano, our model achieves $79.710$ beating the Random Forest model presented in \citet{imperial-etal-2022-baseline} with a score of $57.485$ with both models utilizing the same Cebuano dataset. Lastly, as there are no automated readability assessment models yet for Bikol, we report a baseline accuracy of $79.328$, which is achieved using a model with a combination of traditional features (extended with \textsc{CrossNGO}) and mBERT embeddings extracted from data in all three Philippine languages.

\begin{table}[!htbp]
\small
\centering
\begin{tabular}{@{}lccc@{}}
\toprule
\textbf{Model} & \textbf{TGL} & \textbf{BCL} & \textbf{CEB} \\
\midrule
\textbf{TGL}     & 0.420          & 0.500          & 0.333          \\
\textbf{BCL}     & 0.420          & 0.633          & 0.575          \\
\textbf{CEB}     & \textbf{0.520} & 0.500          & \textbf{0.697} \\\midrule
\textbf{TGL+BCL} & 0.440          & 0.566          & 0.469          \\
\textbf{BCL+CEB} & 0.400          & \textbf{0.637} & 0.666          \\
\textbf{CEB+TGL} & 0.480          & 0.500          & 0.590          \\\midrule
\textbf{*ALL}    & 0.460          & 0.633          & 0.636          \\
\bottomrule
\end{tabular}
\caption{The accuracy scores of the conventional fine-tuning strategy applied to LLMs for various methods of cross-lingual ARA using the same uncased mBERT model for the extraction of embeddings.}
\label{tab:finetune_mbert}
\end{table}

\subsection{Conventional Fine-Tuning of mBERT Underperforms for Low Resource Cross-Lingual ARA}
While the main focus of our work is on using traditional machine learning models with Random Forest, we explore if the standard approach for fine-tuning LLMs such as mBERT can produce comparable performance. We use the same uncased mBERT model as presented in Section \ref{sec:setup}.

Table~\ref{tab:finetune_mbert} shows the performance of singular, pairwise, and full cross-lingual setups formatted similarly to Table~\ref{tab:crossmodelling_result}. These results confirm the findings of \citet{ibanez2022applicability}, who have applied a similar setup to monolingual Tagalog ARA using a Tagalog BERT model. Judging by their results, the conventional fine-tuning approach proved to be inferior to the traditional way of extracting linguistic features from text and training a machine learning model like SVM or Random Forest. For this study, the highest-performing setups for Tagalog and Cebuano use Cebuano data only, and that for Bikol uses the combined Cebuano + Bikol datasets. None of the fine-tuned models outperform those presented in Table~\ref{tab:crossmodelling_result} using combinations of traditional features and \textsc{CrossNGO}. While previous work in cross-lingual ARA by \citet{lee-vajjala-2022-neural} and \citet{madrazo2020cross} achieved relatively high performance with non-closely related languages using LLMs, we obtain less promising results which we can attribute to: (a) the use of datasets of substantially smaller sizes (a total of $13,786$ documents used in \citet{azpiazu2019multiattentive} and $17,518$ in \citet{lee-vajjala-2022-neural} vs. only $764$ in out study), and (b) lack of diverse data sources since only Wikipedia dumps were used for Tagalog and Cebuano for training the mBERT model.

%%have applied a similar approach to the  English, French, and Spanish datasets from Newsela \cite{xu2015problems}, \citet{madrazo2020cross} with the VikiWiki dataset \cite{azpiazu2019multiattentive} and achieved  high results $> 80\%$ in accuracy. 

\section{Conclusion}
%While readability assessment in low-resource languages is slowly gaining momentum from the research community, most of the approaches used for this task are limited to extracting traditional handcrafted linguistic features as predictors of text complexity. 
In this work, we took a step back from the trend of exploring various technical components of the complex, deep learning models and, instead, focused on studying the potential effectiveness of linguistic characteristics such as mutual intelligibility for ARA in closely related Philippine languages – Tagalog, Bikol, and Cebuano. We implemented three cross-lingual setups to closely study the effects of interaction between the three languages and proposed a new feature utilizing n-gram overlap, \textsc{CrossNGO}, which is specially developed for cross-lingual ARA using closely related languages. Our results show that: (a) using \textsc{CrossNGO} combined with handcrafted features achieves significantly higher performance than using mBERT embeddings, (b) the inclusion of another closely related Philippine language reduces model confusion, and (c) using the conventional fine-tuning for LLMs like mBERT in this setup still does not outperform models with traditional features. Consequently, we come to the conclusion that using languages with high intelligibility is more suited for cross-lingual ARA. This is demonstrated in experiments with English added as an example of a non-related language, in which we do not achieve a substantial increase in performances for Tagalog, Cebuano, and Bikol.

Our results agree with the findings of previous studies in cross-lingual ARA such as those of \citet{madrazo2020cross} using English, Spanish, Basque, Italian, French, Catalan, and \citet{weiss-etal-2021-using} using English and German, that also showed that the inclusion of additional language data can improve ARA results on other languages. However, our work is primarily motivated by the degree of language relatedness: we show that better results can be achieved for ARA in low-resource languages if we use closely related languages rather than any language, including non-related ones like English. Our  study also provides an encouragement for researchers to consider approaches grounded in linguistic theories which can potentially be used to improve the performance in NLP tasks rather than always resorting to models that are expensive to train and hard to interpret.

\section{Limitations}
We discuss some limitations of our current work which can be further explored in the future. \newline

\noindent \textbf{On Data Format}. We specifically use fictional short stories as our primary data for the study since we require gold standard labels for this document classification task. Moreover, fictional short stories are easier to find as they often come with a specified grade level compared to other types of literary texts such as magazines or web articles written in any of the three Philippine languages. We do not claim that our models are able to generalize on these other types of literary materials or on other types of closely related language pairs unless a full study is conducted which is outside the scope of this work.
\newline

\noindent \textbf{On Handcrafted Features}. We were only able to use traditional handcrafted features covering count-based predictors such as sentence or word count and syllable pattern-based features for training the Random Forest models. We did not extract other feature sets one may find in the previous work on English such as lexical density or discourse-based features since such features require NLP tools that are able to extract POS, named entities, relations, and discourse patterns that do not yet exist for all three Philippine languages used in this study. The work of \citet{imperial-ong-2021-microscope} covered a small set of lexical features such as \textit{type–token ratio} and \textit{compound word density} for readability assessment in Tagalog. Still, we cannot use this approach since all languages would need to have the same number of features as is a standard practice in model training. \newline

\noindent \textbf{On Model Training}. Our choice of the Random Forest algorithm for training the ARA models is based on the substantial amount of previous work supporting the application of this method to low-resource ARA, e.g., to Tagalog and Cebuano in a monolingual setup \cite{imperial2020exploring,imperial2021diverse,imperial-2021-bert,imperial-etal-2022-baseline}, where it achieved better results than other algorithms such as SVM or Logistic Regression. One can consider these algorithms for comparison but the analysis of each ARA model trained with various algorithms to the same level of depth and focus that we have given to the Random Forest classifier in the present study would require a considerable amount of time as well as a higher page limit. \newline

\noindent \textbf{On Current Measures of Mutual Intelligibility}. The majority of existing literature in linguistics, specifically on the topic of mutual intelligibility in Philippine languages, discusses examples in the context of speech communication. As such, one might claim that Cebuano and Tagalog are \textit{not} mutually intelligible by giving an example where a Tagalog speaker may not fully comprehend (or only recognize a few common words) another speaker if they are talking in Cebuano. While this is certainly true, in this study, we specifically focus on the mutual intelligibility of languages at a word and character level via written texts such as children's fiction books. From this, we see a substantial degree of \textit{closeness} between Tagalog, Cebuano, and Bikol compared to English. Thus, based on our results, we posit that mutual intelligibility may be used as an additional feature (see \textsc{CrossNGO} in Section \ref{sec:setup}) for text-based tasks such as readability assessment. We leave the exploration of our proposed novel feature in the speech communication area to future work.

\section{Ethical Considerations}
We foresee no ethical issues related to the study. 
%The work does not involve human subjects, use of private or sensitive data, or exhaustive training of large language models.

\section*{Acknowledgements}
We thank the anonymous reviewers and area chairs for their constructive and helpful feedback. We also thank the communities and organizations behind the creation of open-source datasets in Philippine languages used in this research: DepED, Adarna House, Bloom Library, Let's Read Asia, SIL, and BookLabs. JMI is supported by the UKRI CDT in Accountable, Responsible, and Transparent AI of the University of Bath and by the Study Grant Program of the National University Philippines.

% Entries for the entire Anthology, followed by custom entries
\bibliography{anthology,custom}
\bibliographystyle{acl_natbib}

%\clearpage
\appendix
\newpage
\section{Appendix}
\label{sec:appendix}

\begin{table}[!htbp]
\centering
\begin{tabular}{@{}ll@{}} \toprule
\bf Hyperparameter & \bf Value                         \\ 
\midrule
batchSize      & 100                           \\
bagSizePercent & 100                           \\
maxDepth       & unlimited                     \\
numIterations  & 100                           \\
numFeatures    & int(log(\#predictors) + 1) \\
seed           & 1                            \\
\bottomrule
\end{tabular}
\caption{Hyperparameter settings for the Random Forest algorithm used for training the models in WEKA. These are default values and the $3.8.6$ version of WEKA would have these already preset.}
\end{table}

\begin{table}[!htbp]
\centering
\begin{tabular}{@{}ll@{}}
\toprule
\bf Hyperparameter & \bf Value \\ \midrule
max seq length & 300   \\
batch size     & 8     \\
dropout        & 0.01  \\
optimizer      & Adam  \\
activation     & ReLu  \\
layer count & 1 (768 x 256) \\
loss           & Negative Log Likelihood \\
learning rate  & 0.002  \\
epochs         & 50    \\ \bottomrule
\end{tabular}
\caption{Hyperparameter settings for the mBERT model used for fine-tuning. Please  refer to \citet{ibanez2022applicability} for more information on these values.}
\end{table}

\begin{table*}[!htbp]
\centering
\begin{tabular}{@{}cccccc@{}}
\toprule
\multicolumn{2}{c}{\bf TGL}                                & \multicolumn{2}{c}{\bf CEB}                                & \multicolumn{2}{c}{\bf BCL}                                \\\midrule
\multicolumn{1}{c}{\bf bigram} & \multicolumn{1}{c}{\bf count} & \multicolumn{1}{c}{\bf bigram} & \multicolumn{1}{c}{\bf count} & \multicolumn{1}{c}{\bf bigram} & \multicolumn{1}{c}{\bf count} \\
\midrule
ng & 43215 & an & 15636 & an & 12562 \\
an & 39268 & ng & 14451 & na & 7315  \\
na & 22041 & sa & 8311  & ng & 6754  \\
in & 18449 & na & 7167  & in & 6138  \\
ma & 16501 & ga & 6714  & sa & 5753  \\
sa & 16037 & ka & 5951  & ka & 5176  \\
la & 15283 & la & 5638  & ag & 4558  \\
ka & 14263 & ma & 4889  & ma & 4452  \\
ag & 12386 & ni & 4701  & on & 3490  \\
at & 12380 & ta & 4692  & ga & 3462  \\
pa & 12171 & in & 4591  & pa & 3453  \\
al & 11521 & pa & 4333  & ni & 3416  \\
ga & 10818 & ag & 4247  & ak & 3291  \\
ay & 10771 & on & 4113  & ar & 3012  \\
ak & 10271 & ay & 3799  & si & 2957  \\
ni & 9814  & si & 3636  & da & 2920  \\
ta & 9738  & ya & 3603  & ya & 2886  \\
si & 9126  & al & 3406  & ta & 2796  \\
ya & 8724  & at & 3150  & la & 2676  \\
on & 8288  & ba & 3099  & al & 2658  \\
ba & 7402  & ak & 3062  & ba & 2613  \\
it & 7288  & ha & 2729  & ra & 2518  \\
am & 6667  & iy & 2634  & as & 2447  \\
iy & 6339  & ug & 2531  & at & 2315  \\
as & 6210  & il & 2511  & ay & 2187  \\
ko & 5928  & un & 2502  & ab & 1893  \\
ha & 5885  & gi & 2460  & ai & 1843  \\
il & 5857  & li & 2413  & ko & 1840  \\
ar & 5848  & am & 2327  & ha & 1763  \\
li & 5696  & ah & 2251  & li & 1697  \\
ap & 5190  & it & 2059  & ad & 1679  \\
ab & 5000  & ad & 1834  & ro & 1574  \\
ra & 4867  & as & 1801  & am & 1544  \\
da & 4777  & da & 1793  & un & 1316  \\
aw & 4598  & us & 1781  & ti & 1293  \\
ti & 4577  & ko & 1771  & nd & 1202  \\
wa & 4572  & to & 1770  & ap & 1172  \\
ah & 4410  & aw & 1767  & mg & 1165  \\
um & 4391  & ab & 1690  & ah & 1164  \\
bi & 4382  & yo & 1667  & it & 1160  \\
is & 4286  & ki & 1615  & bi & 1146  \\
to & 4248  & hi & 1589  & ku & 1140  \\
mi & 4179  & ap & 1516  & aw & 1139  \\
un & 4168  & mg & 1504  & wa & 1086  \\
\bottomrule
\end{tabular}
\caption{Full list of the top $25\%$ bigrams extracted from the Tagalog, Cebuano, and Bikol datasets. The same list is used for calculating overlap via \textsc{CrossNGO}.}
\end{table*}

\begin{table*}[!htbp]
\centering
\begin{tabular}{@{}cccccc@{}}
\toprule
\multicolumn{2}{c}{\bf TGL} &
  \multicolumn{2}{c}{\bf CEB} &
  \multicolumn{2}{c}{\bf BCL} \\ \midrule
\multicolumn{1}{c}{\bf trigram} &
  \multicolumn{1}{c}{\bf count} &
  \multicolumn{1}{c}{\bf trigram} &
  \multicolumn{1}{c}{\bf count} &
  \multicolumn{1}{c}{\bf trigram} &
  \multicolumn{1}{c}{\bf count} \\ \midrule
ang & 22650 & ang & 7941 & ang & 3350 \\
ala & 6120  & nga & 3283 & nag & 1721 \\
ing & 5456  & iya & 2547 & kan & 1518 \\
ong & 5036  & ing & 1697 & aka & 1507 \\
iya & 4761  & ala & 1534 & ing & 1434 \\
lan & 3880  & mga & 1479 & nin & 1389 \\
ina & 3481  & ila & 1474 & ong & 1374 \\
aka & 3266  & ana & 1395 & ara & 1210 \\
nan & 3151  & lan & 1317 & mga & 1164 \\
ama & 3021  & ong & 1315 & man & 1103 \\
ara & 3007  & ata & 1306 & yan & 979  \\
ata & 2976  & usa & 1286 & sin & 947  \\
ila & 2965  & tan & 1276 & ala & 940  \\
mga & 2867  & yan & 1172 & iya & 928  \\
nag & 2797  & han & 1139 & asi & 897  \\
niy & 2795  & ali & 1061 & sai & 853  \\
pag & 2793  & nag & 1043 & aba & 835  \\
yan & 2757  & pag & 982  & ina & 833  \\
apa & 2716  & aka & 975  & aga & 824  \\
aga & 2694  & ayo & 933  & ini & 816  \\
ali & 2622  & aha & 931  & mag & 812  \\
man & 2574  & nan & 928  & aro & 730  \\
aha & 2450  & siy & 916  & ako & 730  \\
uma & 2412  & ako & 868  & gan & 718  \\
aki & 2376  & pan & 863  & par & 705  \\
nga & 2281  & ama & 847  & nbs & 702  \\
mag & 2269  & man & 831  & bsp & 702  \\
aba & 2253  & ini & 830  & ata & 683  \\
awa & 2249  & ita & 827  & nga & 683  \\
kan & 2219  & una & 811  & pag & 639  \\
tin & 2208  & ina & 763  & ati & 605  \\
asa & 2142  & aba & 758  & lan & 582  \\
ako & 2130  & kin & 744  & ion & 576  \\
hin & 2119  & nak & 727  & nda & 574  \\
ito & 2033  & ung & 718  & lin & 569  \\
aya & 2000  & kan & 716  & sak & 567  \\
ana & 1993  & san & 700  & ano & 553  \\
gan & 1973  & nah & 700  & ban & 547  \\
ami & 1934  & ngo & 679  & ind & 538  \\
san & 1913  & kat & 675  & ron & 530  \\
nak & 1896  & gan & 665  & apa & 527  \\
abi & 1878  & ula & 636  & ana & 526  \\
tan & 1844  & ano & 626  & ili & 524  \\
siy & 1835  & uot & 611  & ent & 508  \\
ani & 1773  & ahi & 605  & ada & 502  \\ \bottomrule

\end{tabular}
\caption{Full list of the top $25\%$ trigrams extracted from the Tagalog, Cebuano, and Bikol datasets. The same list is used for calculating overlap via \textsc{CrossNGO}.}
\end{table*}

%\begin{table*}[!htbp]
%\centering
%\begin{tabular}{@{}ll@{}} \toprule
%\bf Hyperparameter & \bf Value                         \\ 
%\midrule
%batchSize      & 100                           \\
%bagSizePercent & 100                           \\
%maxDepth       & unlimited                     \\
%numIterations  & 100                           \\
%numFeatures    & int(log(\#predictors) + 1) \\
%seed           & 1                            \\
%\bottomrule
%\end{tabular}
%\caption{Full list of specifications for the Random Forest algorithm used for training the models in WEKA. These are default values and using the 3.8.6 version of WEKA would have these already preset.}
%\end{table*}

%\begin{table*}[!htbp]
%\centering
%\begin{tabular}{@{}ll@{}}
%\toprule
%\bf Hyperparameter & \bf Value \\ \midrule
%max seq length & 300   \\
%batch size     & 8     \\
%dropout        & 0.01  \\
%optimizer      & Adam  \\
%activation     & ReLu  \\
%layer count & 1 (768 x 256) \\
%loss           & Negative Log Likelihood \\
%learning rate  & 0.002  \\
%epochs         & 50    \\ \bottomrule
%\end{tabular}
%\caption{Full list of specifications for the mBERT model used for finetuning. We referred from the work of \citet{ibanez2022applicability} for these values.}
%\end{table*}

\end{document}